\documentclass[journal]{IEEEtran}
\usepackage{times}

\usepackage[numbers]{natbib}
\usepackage{multicol}
\usepackage[bookmarks=true]{hyperref}
\usepackage{algorithm}
\usepackage{bm}
\usepackage{amssymb}
\usepackage{amsmath}
\usepackage{graphicx}
\usepackage{epstopdf}
\usepackage{amsfonts}%
\usepackage{indentfirst}
\usepackage{array}
\usepackage{color}
\usepackage{wrapfig}
\usepackage{booktabs}
\usepackage{subcaption}

\newtheorem{problem}{\textbf{Problem}}
\newtheorem{definition}{\rm\textbf{Definition}}
\newtheorem{theorem}{\rm\textbf{Theorem}}

\begin{document}

\title{Differentiable Control Barrier Functions for Vision-based End-to-End Autonomous Driving}

\author{Wei Xiao$^*$, Tsun-Hsuan Wang$^*$, Makram Chahine, Alexander Amini, Ramin Hasani and Daniela Rus
\thanks{$^*$The authors are with equal contributions.}
\thanks{The authors are with the Computer Science and Artificial Intelligence Lab, Massachusetts Institute of Technology \texttt{{\small \{weixy, tsunw, chahine, amini, rhasani, rus\}@mit.edu}}}
}



%

\maketitle

\begin{abstract}
Guaranteeing safety of perception-based learning systems is challenging due to the absence of ground-truth state information unlike in state-aware control scenarios. In this paper, we introduce a safety guaranteed learning framework for vision-based end-to-end autonomous driving. To this end, we design a learning system equipped with differentiable control barrier functions (dCBFs) that is trained end-to-end by gradient descent. Our models are composed of conventional neural network architectures and dCBFs. They are interpretable at scale, achieve great test performance under limited training data, and are safety guaranteed in a series of autonomous driving scenarios such as lane keeping and obstacle avoidance. 
We evaluated our framework in a sim-to-real environment, and tested on a real autonomous car, achieving safe lane following and obstacle avoidance via Augmented Reality (AR) and real parked vehicles.
\end{abstract}

\IEEEpeerreviewmaketitle

\section{Introduction}

Neural networks are powerful tools for learning relevant representations in complex scenarios. However, applying such learning systems in decision and control problems such as autonomous driving is significantly hindered by the absence of safety assurance.  This is due to the learning systems being a black box that makes it complex to perform root cause analysis. Thus, a single mistake made by a learned neural controller can potentially lead to catastrophic outcomes. Ensuring safety (e.g., providing guarantees that a self-driving car will never collide with obstacles) of a learning system is therefore very important.  
Nevertheless, as the dimensionality of the observations and action space for a real-world problem increases, defining safety criteria and guarantees for learning systems increases dramatically. Take for instance a vision-based autonomous driving algorithm; even identifying the contributions of every single pixel to making driving decisions in every scenario is computationally intractable. In this case, how can we ensure the safety of the learning system?

In this paper we leverage theoretical results in differentiable control barrier functions (dCBF) to equip end-to-end vision-based learning systems with safety guarantees. Barrier functions (BFs) have been widely used in optimization formulations to guarantee the satisfaction of some constraints \cite{Khalil2002}, and have recently been extended to Lyapunov-like functions \cite{Tee2009,Wieland2007}. They have been employed to prove set invariance \cite{Aubin2009,Prajna2007,Wisniewski2013} and regulate multi-objective control \cite{Panagou2013}. Control Barrier Functions (CBFs) are extensions of BFs for control systems, and are used to map a constraint defined over system states onto a constraint on the control input \cite{ames2014control}. The satisfaction of the control constraint thus implies the satisfaction of the original safety constraint. Recently, it has been shown that by optimizing a quadratic cost and satisfying state and control
constraints, CBFs can be used to form
quadratic programs (QPs) \cite{nguyen2016exponential}, \cite{ames2014control}, 
\cite{wang2018safe} which can be solved in real time. Prior work also introduced differentiable CBFs as novel CBF formulations whose parameters are trainable \cite{Xiao2020CDC}, \cite{Parwana2021}. They have been incorporated into differentiable QPs \cite{Amos2017} which can in turn be combined with learning systems \cite{Xiao2021bnet}.

\begin{figure}[t]
	\centering
	\includegraphics[scale=1.2]{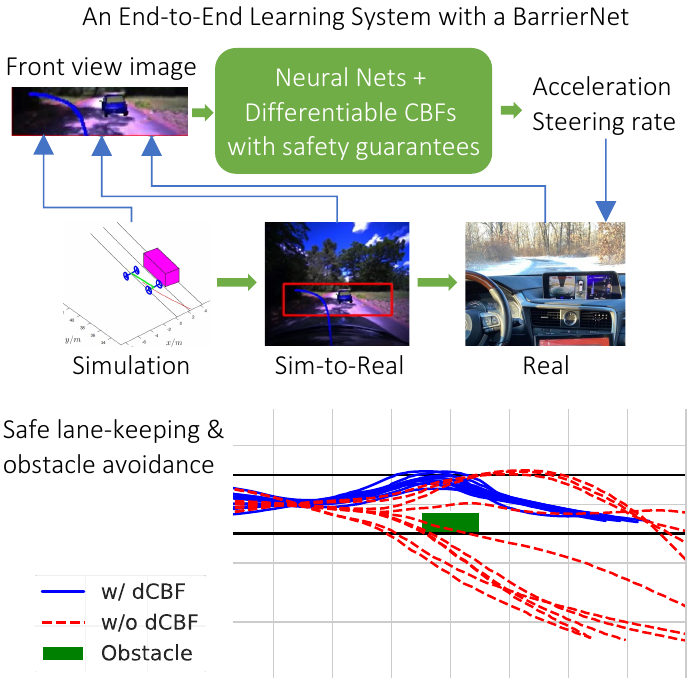}
	\caption{Vision-based end-to-end autonomous driving with differentiable CBFs in a BarrierNet. Lane keeping and collision avoidance are guaranteed.  }	
	\label{fig:Nctrl}
\end{figure}

dCBFs are introduced to mitigate the conservativeness of CBFs for safety guarantees. In CBFs, a system tends to stay far away from the unsafe set boundary, and thus may deviate largely from a desired trajectory. A BarrierNet \cite{Xiao2021bnet} is constructed by incorporating dCBFs into a differentiable QP. In a BarrierNet, all QP parameters, including those originating in CBFs, are trainable. Thus, we obtain a ubiquitous safety guaranteed barrier layer that can be combined with any learning system \citep{hochreiter1997long,he2016deep,vaswani2017attention,lechner2020neural,hasani2021liquid,lechner2021mixed,hasani2021closed}. BarrierNet addresses the conservativeness of CBFs and allows the safety constraints of a neural controller to be adaptable to changing environments. 

Although BarrierNets show promise in guaranteeing safety for shallow neural controllers, they have not been studied in the context of high-dimensional observation spaces such as vision-based control. In this paper, we investigate the effectiveness and conditions for using BarrierNets in end-to-end vision-based autonomous driving scenarios (see Fig. \ref{fig:Nctrl}). In this example, a neural network receives a camera input stream and outputs acceleration and steering rate commands to navigate the vehicle along the center of the driving lane, while avoiding obstacles. The system and environment observations are inputs to the upstream network whose outputs serve as arguments to the BarrierNet layer. Finally, BarrierNet outputs the controls that guarantee collision avoidance.

\noindent \textbf{Contributions:} $(i)$ We present a new end-to-end learning pipeline composed of conventional deep learning models and BarrierNets for vision-based end-to-end autonomous driving to achieve safe maneuvering. $(ii)$ Our algorithm is multi-level interpretable and can achieve good test performance under limited training data. 
$(iii)$ We design a vision-based state estimation module within our pipeline and study how BarrierNet works in the absence of ground truth information
$(iv)$ We train and verify our proposed framework in a sim-to-real environment; we also deploy our model on a full-scale autonomous vehicle for safe lane following and obstacle-avoidance via Augmented Reality (AR) and physically parked vehicles.

\section{Related Work}
\noindent \textbf{Safety-critical control and learning.} A large body of work studied CBF-based safety guarantees \cite{ames2014control, nguyen2016exponential,wang2018safe,choi2020reinforcement,taylor2020learning}. Many existing works \cite{ames2014control,nguyen2016exponential,Yang2019}
combine CBFs for systems with quadratic costs to form optimization problems. Time is discretized and an optimization problem with
constraints given by the CBFs is solved at each time step. Replacing CBFs by High Order Control Barrie Functions (HOCBFs) allows us to handle constraints with arbitrary relative degree \cite{Xiao2019}. The common observation is that CBFs tend to make the system's behavior excessively conservative if they are not properly defined. This conservativeness is usually characterized by how close the system state can (but not necessarily) stay to the unsafe set boundary or a desired reference trajectory in obstacle-clustered environment. 

In order to address the conservativeness of the CBF method, the work in \cite{Xiao2020CDC} proposed to parameterize the definition of CBFs, and use machine learning methods to learn CBF parameters for a certain type of unsafe sets such that safety is guaranteed without control being excessively conservative. This form of CBFs is differentiable. In \cite{Parwana2021} learning CBF parameters using a differential policy over a time horizon is proposed. Adaptive CBFs (AdaCBFs) \cite{Xiao2021TAC1} allow for time-varying CBF parameters. It has been shown that the satisfaction of the AdaCBF constraint is a necessary and sufficient condition for the satisfaction of the original safety constraint. All these related works require the design of proper policies and models, which is tedious and non-trivial. In contrast, we propose to use an end-to-end differentiable framework to learn neural network controllers for safety-critical systems.

At the intersection of CBFs and learning, supervised learning techniques have been proposed to learn safe set definitions from demonstrations \cite{Robey2020} which can be then enforced by CBFs. The authors in \cite{taylor2020learning} used data to learn system dynamics for CBFs.  In \cite{Yaghoubi2020}, neural network controllers are trained using CBFs in the presence of disturbances. These prior works focus on learning safe sets and dynamics, whereas we focus on the design of environment dependent and trainable CBFs.

\noindent \textbf{Differentiable optimization-based safety frameworks.} Recent advances in differentiable optimization methods show promise for safety guaranteed neural network controllers \cite{Amos2017, pereira2020, Amos2018, Xiao2021bnet}. In \cite{Amos2017}, a differentiable quadratic program (QP) layer, called OptNet, was introduced. OptNet with CBFs has been used in neural networks as a filter for safe controls \cite{pereira2020}, but OptNet is not trainable, thus, potentially limiting the system's learning performance.  In \cite{Deshmukh2019,Jin2020,Zhao2021,gruenbacher2020verification,gruenbacher2021gotube}, safety guaranteed neural network controllers have been learned through verification-in-the-loop training. A safe neural network filter has been proposed in \cite{Ferlez2020} for a specific vehicle model using verification methods. The verification approaches cannot ensure coverage of the entire state space. They are offline methods, unable to adapt to environment changes (such as varying size of the unsafe sets) \cite{li2021comparison}. By comparison, the BarrierNet in \cite{Xiao2021bnet} incorporates differentiable CBFs into neural network controllers. 
In this work, we address the challenges of using BarrierNet to achieve vision-based end-to-end autonomous driving.

\noindent\textbf{Sim-to-real end-to-end autonomous driving.} 
Many works have demonstrated the ability to learn end-to-end (perception-to-control) driving policies directly from real-world perception data using imitation learning (IL)~\cite{pomerleau1989alvinn, bojarski2016end}. Such approaches have been demonstrated to be successfully deployed in both real-world offline passive datasets ~\cite{xu2017end, toromanoff2018end, amini2019variational} as well as real closed-loop control test environments~\cite{patel2017sensor, codevilla2018end, lechner2020neural, hawke2020urban}. 

However, these works have focused on ``reactive'' scenarios such as lane-keeping~\cite{bojarski2016end, xu2017end, toromanoff2018end}, lane-changing~\cite{bojarski2020nvidia}, and navigation~\cite{patel2017sensor, codevilla2018end, amini2019variational, hawke2020urban}, but lack the ability to plan a path around other objects in the scene due to significantly larger data requirements of IL to achieve sufficient coverage of the testing distribution. Simulation has emerged as a viable candidate to overcome this challenge and render a continuum of scenarios for learning in the presence of other objects and agents in the environment. Works that learn object avoidance in simulation have leveraged both imitation learning~\cite{xiao2019multimodal} as well as reinforcement learning~\cite{bohez2017sensor,lechner2019designing,brunnbauer2021latent,zeng2022dreaming,hasani2020natural} but often face limited to no deployment capabilities in reality due to large sim-to-real gaps present in model-based simulation. In this work, we leverage recent advances in data-driven simulation~\cite{amini2020learning, li2019aads, wang2021learning, amini2021vista} to overcome the sim-to-real gap to learn robust end-to-end controllers capable of transferring to real scenarios with other agents.

\section{Background}
\label{sec:bkg}
\noindent In this section, we briefly introduce control barrier functions (CBF) and refer interested readers to \citep{ames2014control} for detailed formulations. Intuitively, CBFs are a means to translate state constraints to control constraints under affine dynamics. The controls that satisfy those constraints can be efficiently solved for by formulating a quadratic program. We start with the definition of class $\mathcal{K}$ functions.

\begin{definition}
	\label{def:classk} (\textit{Class $\mathcal{K}$ function} \cite{Khalil2002}) A
	continuous function $\alpha:[0,a)\rightarrow[0,\infty), a > 0$ is said to
	belong to class $\mathcal{K}$ if it is strictly increasing and $\alpha(0)=0$. A continuous function $\beta:\mathbb{R}\rightarrow\mathbb{R}$ is said to belong to extended class $\mathcal{K}$ if it is strictly increasing and $\beta(0)=0$.
\end{definition}

\noindent Consider an affine control system of the form
\begin{equation}
\dot{\bm{x}}=f(\bm x)+g(\bm x)\bm u \label{eqn:affine}%
\end{equation}
where $\bm x\in\mathbb{R}^{n}$, $f:\mathbb{R}^{n}\rightarrow\mathbb{R}^{n}$
and $g:\mathbb{R}^{n}\rightarrow\mathbb{R}^{n\times q}$ are {locally}
Lipschitz, and $\bm u\in U\subset\mathbb{R}^{q}$, where $U$ denotes a control constraint set.

\begin{definition}
	\label{def:forwardinv} A set $C\subset\mathbb{R}^{n}$ is forward invariant for
	system (\ref{eqn:affine}) if its solutions for some $\bm u\in U$ starting at any $\bm x(0) \in C$ satisfy $\bm x(t)\in C,$ $\forall t\geq0$.
\end{definition}

\begin{definition}
	\label{def:relative} (\textit{Relative degree}) The relative degree of a
	(sufficiently many times) differentiable function $b:\mathbb{R}^{n}%
	\rightarrow\mathbb{R}$ with respect to system (\ref{eqn:affine}) is the number
	of times it needs to be differentiated along its dynamics until the control
	$\bm u$ explicitly shows in the corresponding derivative.
\end{definition}

\noindent Since function $b$ is used to define a (safety) constraint $b(\bm
x)\geq0$, we will also refer to the relative degree of $b$ as the relative
degree of the constraint. For a constraint $b(\bm x)\geq0$ with relative
degree $m$, $b:\mathbb{R}^{n}\rightarrow\mathbb{R}$, and $\psi_{0}(\bm
x):=b(\bm x)$, we define a sequence of functions $\psi_{i}:\mathbb{R}%
^{n}\rightarrow\mathbb{R},i\in\{1,\dots,m\}$:
\begin{equation}
\begin{aligned} \psi_i(\bm x) := \dot \psi_{i-1}(\bm x) + \alpha_i(\psi_{i-1}(\bm x)),\quad i\in\{1,\dots,m\}, \end{aligned} \label{eqn:functions}%
\end{equation}
where $\alpha_{i}(\cdot),i\in\{1,\dots,m\}$ denotes a $(m-i)^{th}$ order
differentiable class $\mathcal{K}$ function.

We further define a sequence of sets $C_{i}, i\in\{1,\dots,m\}$ associated
with (\ref{eqn:functions}) in the form:
\begin{equation}
\label{eqn:sets}\begin{aligned} C_i := \{\bm x \in \mathbb{R}^n: \psi_{i-1}(\bm x) \geq 0\}, \quad i\in\{1,\dots,m\}. \end{aligned}
\end{equation}

\begin{definition}
	\label{def:hocbf} (\textit{High Order Control Barrier Function (HOCBF)}
	\cite{Xiao2019}) Let $C_{1}, \dots, C_{m}$ be defined by (\ref{eqn:sets}%
	) and $\psi_{1}(\bm x), \dots, \psi_{m}(\bm x)$ be defined by
	(\ref{eqn:functions}). A function $b: \mathbb{R}^{n}\rightarrow\mathbb{R}$ is
	a High Order Control Barrier Function (HOCBF) of relative degree $m$ for
	system (\ref{eqn:affine}) if there exist $(m-i)^{th}$ order differentiable
	class $\mathcal{K}$ functions $\alpha_{i},i\in\{1,\dots,m-1\}$ and a class
	$\mathcal{K}$ function $\alpha_{m}$ such that
	\begin{equation}
	\label{eqn:constraint}\begin{aligned} 
	\sup_{\bm u\in U}\Big[L_f^{m}b(\bm x) + [L_gL_f^{m-1}b(\bm x)]\bm u + O(b(\bm x)) \\+ \alpha_m(\psi_{m-1}(\bm x))\Big] \geq 0, \end{aligned}
	\end{equation}
	for all $\bm x\in C_{1} \cap,\dots, \cap C_{m}$. In
	(\ref{eqn:constraint}), $L_{f}^{m}$ ($L_{g}$) denotes Lie derivatives along
	$f$ ($g$) $m$ (one) times, and $O(b(\bm x)) = \sum_{i = 1}^{m-1}L_f^i(\alpha_{m-i}\circ\psi_{m-i-1})(\bm x).$ {Further, $b(\bm x)$ is such that $L_gL_f^{m-1}b(\bm x)\ne 0$ on the boundary of the set $C_{1} \cap,\dots, \cap C_{m}$.}
\end{definition}

Note that by setting $m = 1$ in a HOCBF, we can get a relative-degree-one CBF constraint:
\begin{equation}
    L_fb(\bm x) + L_gb(\bm x) + \alpha_1(b(\bm x)) \geq 0.
\end{equation}

\begin{theorem}
	\label{thm:hocbf} (\citep{Xiao2019}) Given a HOCBF $b(\bm x)$ from Def.
	\ref{def:hocbf} with the associated sets $C_{1}, \dots, C_{m}$ defined
	by (\ref{eqn:sets}), if $\bm x(0) \in C_{1} \cap,\dots,\cap C_{m}$,
	then any Lipschitz continuous controller $\bm u(t)$ that satisfies
	the constraint in (\ref{eqn:constraint}), $\forall t\geq0$ renders $C_{1}\cap,\dots,
	\cap C_{m}$ forward invariant for system (\ref{eqn:affine}).
\end{theorem}

Combining CBFs with a quadratic cost $\int_{t_0}^{t_f} \bm u^T(t)H\bm u(t)$, where $H$ is positive definite, we can formulate CBF-based QPs:
\begin{equation} \label{eqn:obj}
\bm u^*(t) = \arg\min_{\bm u(t)} \frac{1}{2}\bm u(t)^TH\bm u(t)
\end{equation}
s.t.
$$
\begin{aligned} 
	L_f^{m}b(\bm x) + [L_gL_f^{m-1}b(\bm x)]\bm u \!+\! O(b(\bm x)) + \alpha_m(\psi_{m-1}(\bm x)) \geq 0 \end{aligned}
$$
$$
\bm u\in U, \quad t = k\Delta t + t_0,
$$
\section{Problem Formulation}
\noindent We now formally define learning of safety-critical control for autonomous driving. 

\begin{problem}\label{prob:1}
Given (i) front-view RGB camera images of the vehicle, (ii) a nominal controller $k^\star(\bm x)=\bm u^\star$ (such as a model predictive controller) (iii) vehicle dynamics in the form of (\ref{eqn:affine}),  (iv) a set of safety constraints $b_j(\bm x) \geq 0, j\in S$ (where $b_j$ is continuously differentiable). Typical safety constraints include obstacle avoidance and lane keeping. (v) control bounds $\bm u_{min}\leq\bm u\leq\bm u_{max}$ of the vehicle, and (vi) a neural network controller $k(\bm x | \theta)=\bm u$ parameterized by $\theta$, our goal is to take (i) as input and find the optimal parameters  

\begin{equation}
    \theta^\star = \underset{\theta}{\arg\min}\mathbb{E}_{\bm x}[l(k^\star(\bm x), k(\bm x|\theta))]
\end{equation}

\noindent while guaranteeing the satisfaction of the safety constraints in (iv) and control bounds in (v). $\mathbb{E}(\cdot)$ is the expectation and $l(\cdot, \cdot)$ denotes a similarity measure. Note that state estimation required in (iii) and (iv) is implicitly done in the neural network by inferring from the vision inputs (i).

\end{problem}

\section{Safety-Aware Differentiable Framework}
\label{sec:safety-aware}
In this section, we propose a safety-guaranteed neural network controller for vision-based end-to-end autonomous driving based on BarrierNet. We first study where the two "ends" should be defined in this framework in order to learn a good model based on limited data.

\subsection{Interpretable End-to-End Design}

\noindent \textbf{System's Inputs Setup.} In human-driving, the majority of the information human drivers rely on comes from the front vision view. Therefore, we let the input of the end-to-end architecture be front view images, which contain enough information for executing safe driving.

\noindent \textbf{System's Outputs Setup} At the output end, where BarrierNet is implemented, high-relative-degree control variables (such as acceleration, jerk, steering rate or steering acceleration) are generated for driving the vehicle. The first advantage of using high-relative-degree control variables is to ensure the smoothness of the vehicle states (such as speed), which ensure the vehicle is maneuvered in a smooth manner so that passenger comfort is met. Another advantage is to ensure the controller works with accurate maneuvering due to the physical inertia of the vehicle. If we take vehicle speed as one of the controls, and the controller requires the speed to suddenly change to a large different value, the vehicle powertrain system will fail to respond. In this case the vehicle control becomes inaccurate, affecting the performance and even the safety of the vehicle.

The main challenge in an end-to-end autonomous driving with high-relative-degree control variables is that it requires a very large training data set with high diversity. This is because more vehicle state variables are involved in higher-relative-degree controls, and thus, the vehicle may have different states under the same observation. This will cause confusion in the training process. If the training dataset is not large and diverse enough, the poor generalization of the trained controller (from open-loop to closed-loop, or from sim-to-real) would make the controller fail to achieve its task. Thus, we propose a multi-level interpretable model under limited training data set as shown next.

\begin{figure}[t]
	\centering
	\vspace{-3mm}
	\includegraphics[scale=0.2]{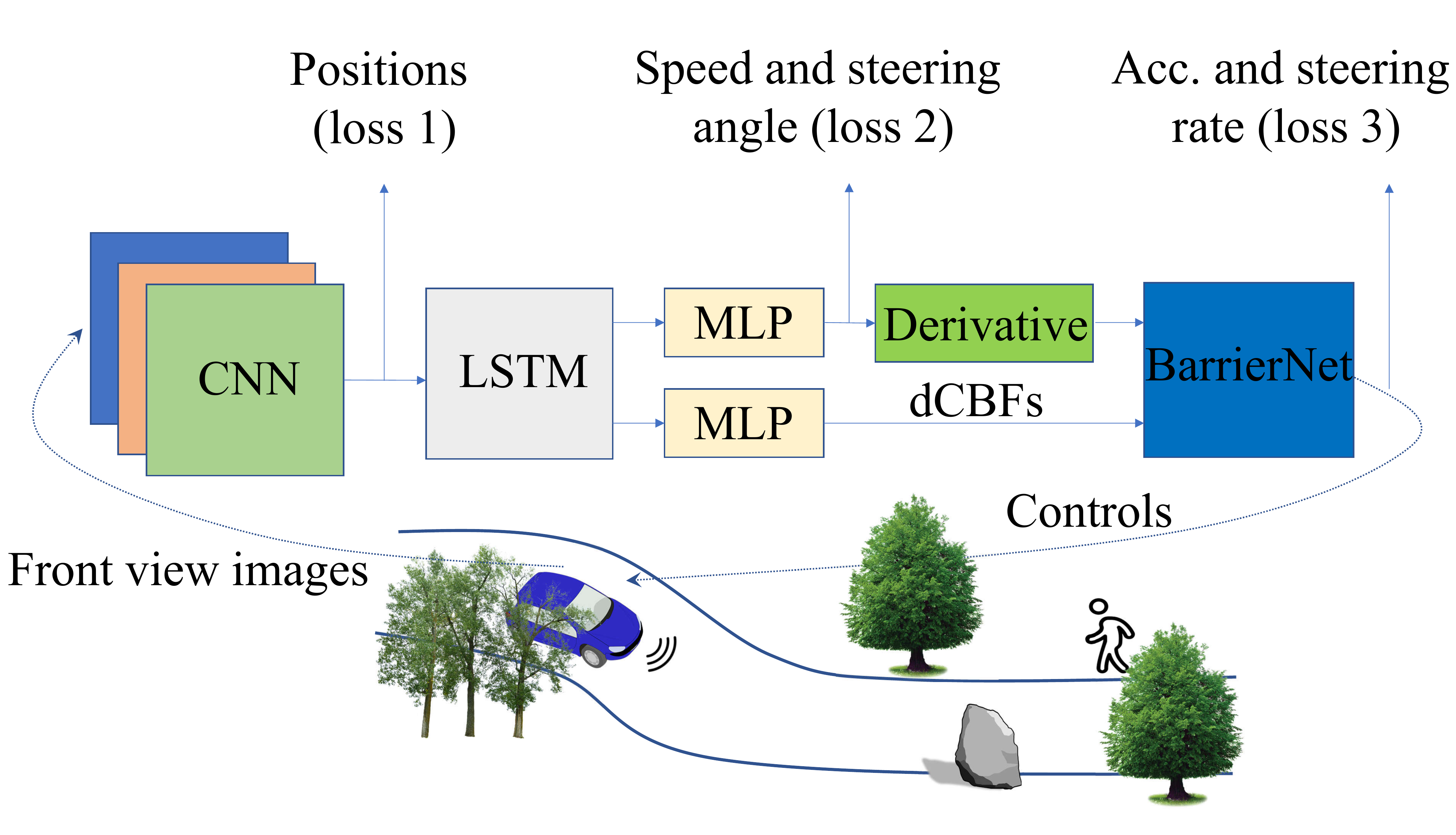}
	\vspace{-2mm}
	\caption{Multi-level interpretable end-to-end autonomous driving framework with differentiable CBFs. The entire pipeline is end-to-end differentiable. Each depth of the model learns different vehicle and environment information. The outputs of BarrierNet are high-relative-degree controls.}
	\label{fig:dcbf}%
	\vspace{-3mm}
\end{figure}

\noindent \textbf{Setting up an interpretable framework.} In order to make the model interpretable, we take training loss outputs at different depths of the model. Following the CNN or LSTM, we may take part of the neurons as the loss outputs for the locations of the vehicle itself and the obstacles. This way, we train neurons at this level to learn position information. In a deeper setting, we may take part of the neurons as another loss outputs for the speed and steering angle of the vehicle, training these neurons to learn speed and steering angle information. By adding derivative layers following these neurons, we get acceleration and steering rate information of the vehicle. The acceleration and steering rate could be taken as reference controls in the BarrierNet which is also trainable, and we take the output of the BarrierNet as the final loss output in the training process. In this framework, different depths of neurons learn different vehicle information, which makes the whole structure consistent with the vehicle physical dynamics. This architecture ensures the neural network model is interpretable \cite{hasani2020interpretable}. We present the whole model structure in Fig. \ref{fig:dcbf}.

\subsection{Differentiable Control Barrier Functions}
\label{ssec:dcbfs}
In this subsection, we briefly introduce BarrierNet \cite{Xiao2021bnet} that incorporates differentiable CBFs, and propose solutions for some of the challenges that arise in autonomous driving with BarrierNet.

\noindent \textbf{Differentiable CBFs} are motivated by the fact that the traditional CBFs can easily make the system overly conservative. In order to address this conservativeness, we multiply the class $\mathcal{K}$ functions in (\ref{eqn:functions}) in the definition of a HOCBF with some observation-dependent functions $p_i(\bm z), i\in\{1,\dots, m\}$:
\begin{equation}
\begin{aligned} \psi_i(\bm x, \bm z) := \dot \psi_{i-1}(\bm x, \bm z) + p_i(\bm z)\alpha_i(\psi_{i-1}(\bm x, \bm z)),\\ i\in\{1,\dots,m\}, \end{aligned} \label{eqn:layers}%
\end{equation}
where $\psi_0(\bm x, \bm z) = b(\bm x)$ and $\bm z \in \mathbb{R}^d$ is the input (such as front view images in autonomous driving) of the neural network ($d\in\mathbb{N}$ is the dimension of the features), $p_i:\mathbb{R}^d\rightarrow \mathbb{R}^{>0}, i\in \{1,\dots,m\}$ are the outputs of the previous layer, where $\mathbb{R}^{>0}$ denotes the set of positive scalars. The above formulation is similar to that of AdaCBF \citep{Xiao2021TAC1}, but, contrary to the latter, is trainable and does not require the design of auxiliary dynamics for $p_i$ (a non-trivial process). To ensure the validity of the above defined CBFs, we require that each $p_i$ be continuously differentiable. Then, we have a differentiable HOCBF as in Def. \ref{def:hocbf} in the form:
\begin{equation}
	\label{eqn:NCBF}\begin{aligned} 
	L_f^{m}b(\bm x) + [L_gL_f^{m-1}b(\bm x)]\bm u \!+\! O(b(\bm x), \bm z) \\+ p_m(\bm z)\alpha_m(\psi_{m-1}(\bm x, \bm z)) \geq 0, \end{aligned}
\end{equation}
Note that it is possible to add additional training parameters to the above class $\mathcal{K}$ functions. For example, we may take the powers as training parameters if the class $\mathcal{K}$ functions are power functions. However, this may decrease the stability of the system as the values of the class $\mathcal{K}$ functions are more sensitive to powers than to coefficients.

\noindent \textbf{BarrierNet.} Eventually, we can incorporate the above softened HOCBFs into differentiable QPs, and obtain a BarrierNet:
\begin{equation} \label{eqn:neuron}
\bm u^*(t) = \arg\min_{\bm u(t)} \frac{1}{2}\bm u(t)^TH(\bm z | \theta_h)\bm u(t) + F^T(\bm z | \theta_f)\bm u(t)
\end{equation}
s.t.
\begin{equation}\label{eqn:10}
\begin{aligned} 
	L_f^{m}b_j(\bm x) + [L_gL_f^{m-1}b_j(\bm x)]\bm u \!+\! O(b_j(\bm x), \bm z| \theta_p) \\+ p_m(\bm z | \theta^m_p)\alpha_m(\psi_{m-1}(\bm x, \bm z|\theta_p)) \geq 0, j\in S \end{aligned}
\end{equation}
$$
\bm u_{min}\leq\bm u\leq\bm u_{max},
$$
$$
t = k\Delta t + t_0,
$$
where $F(\bm z | \theta_f)\in\mathbb{R}^q$ could be interpreted as a reference control (can be the output of previous network layers) and $\theta_h, \theta_f, \theta_p = (\theta_p^1, \dots, \theta_p^m)$ are trainable parameters. $S$ denotes a set of safety constraints including obstacle avoidance and lane keeping. The above differentiable QPs formulate a neuron in BarrierNet. We let both $H(\bm z | \theta_h)$ and $F(\bm z | \theta_f)$ be parameterized and dependent on the network input $\bm z$, but $H$ and $F$ can also be directly trainable parameters that do not depend on the previous layer (i.e., we have $H$ and $F$). The same applies to $p_i, i\in\{1,\dots, m\}$. The trainable parameters are $\theta = \{\theta_h, \theta_f, \theta_p\}$ (or $\theta = \{H, F, p_i, \forall i\in\{1,\dots,m\}\}$ if $H, F$ and $p_i$ do not depend on the previous layer). The solution $\bm u^*$ is the output of the neuron. The BarrierNet layer is differentiable with respect to its parameters \citep{Amos2017}.

\noindent \textbf{Safety with an unknown number of constraints.} One of the challenges in autonomous driving with BarrierNet is that we have to define the exact number of the HOCBFs when designing the BarrierNet layer as it connects with previous layers. However, a vehicle may encounter time-varying number of obstacles (constraints) in a complex environment. In order to address this problem, we proceed as follows.
\begin{figure}[thpb]
	\centering
	\vspace{-3mm}
	\includegraphics[scale=0.28]{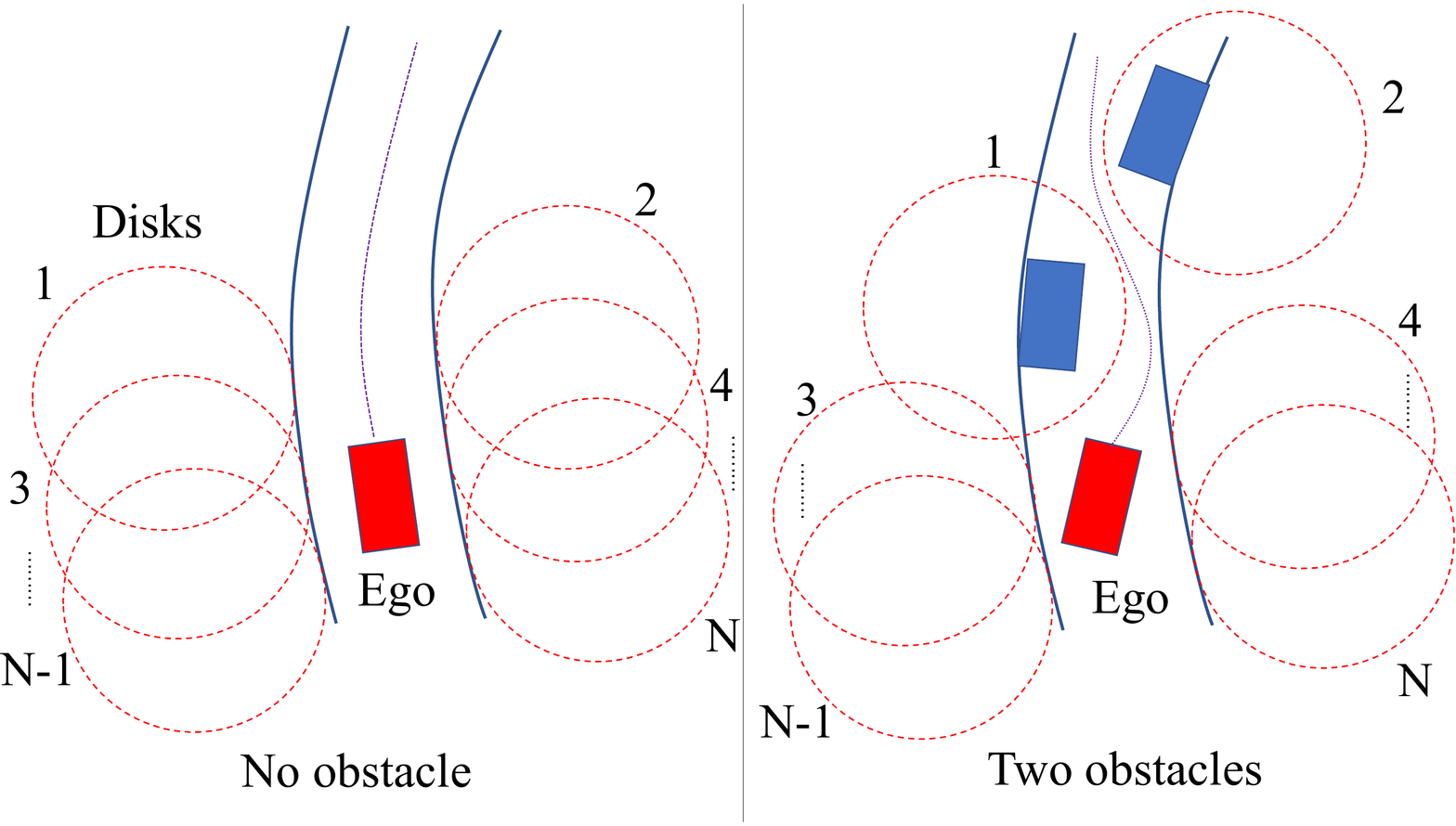}
	\vspace{-7mm}
	\caption{Large disk covering approach for obstacle avoidance. Collision can be avoided if the center of the vehicle never enters the disks (for a correct design of disk locations and sizes). Sorted disks are used to cover corresponding sorted obstacles as they present themselves.}
	\label{fig:cover}%
	\vspace{-3mm}
\end{figure}

Suppose $N\in\mathbb{N}$ denotes the maximum number of obstacles (such as other vehicles) a vehicle may encounter in driving. We cover each of the obstacles with an off-the-center disk, as shown in Fig. \ref{fig:cover}. The deviation direction of the disk depends on the direction of the obstacle with respect to the lane center. In this manner, we may use large disks to cover obstacles while making sure that the ego vehicle will not be overly conservative in driving through. We may use multiple small disks to cover a single obstacle. However, this increases the number of safety constraints required. Another advantage of using a large off-the-center disks is to ensure the smoothness of the vehicle trajectory and to avoid getting stuck in local traps that may appear with small disks. In this setting, we only have $N$ safety constraints, one for each obstacle. We sort them in a specific order in the connection with the previous layer, and enforce them using the above differentiable HOCBFs.

When there are no actual obstacles on the road, as in the case depicted on the left-hand side of Fig. \ref{fig:cover}, we just move the covering disks off the road in which case they play the role of lane keeping. These disks move along the road as the vehicle progresses at the same speed. While the vehicle drives on the road, these disks do not affect its motion as the corresponding HOCBF constraints are not activated. However, if the vehicle is about to leave the road, these constraints can prevent it from doing so due to the safety guarantees of HOCBFs.

When there is one or more obstacles on the road, as in the case depicted on the right-hand side of Fig. \ref{fig:cover}, we first sort the obstacles according to their distance with respect to the ego vehicle. Then, we use the sorted disks to cover the corresponding sorted obstacles. The sorted covering approach can make sure that vehicle may leave the road in order to avoid collision with obstacles. In this setting, although we may have redundant differentiable HOCBFs in terms of obstacle avoidance, these HOCBFs always play an important role in guiding the vehicle, either in lane keeping or obstacle avoidance.

\noindent \textbf{Tackling potential conflicts of HOCBF constraints and control bounds.} Another challenge for BarrierNet is that the differentiable QPs can easily become infeasible to solve during training due to the possible conflict between the HOCBF constraints and the control bounds. In order to address this, we need to require that the nominal control provides control labels that strictly satisfy the safety constraints and control bounds. Then, during training in the BarrierNet layer, we can relax/remove control bounds. After the neural network converges, the differentiable QPs would be feasible when we add control bounds in the testing or implementation. However, there is still possibility that the QP could be infeasible as the BarrierNet may have some inputs that it has never seen before. In order to address this, we can find sufficient conditions of feasibility, as shown in \cite{Xiao2021}. Briefly, this approach finds a feasibility constraint on the state of system and the penalties $p_i(\bm z), i\in\{1,\dots,m\}$, and then enforces this feasibility constraint using another CBF.

\section{Experiments}
In this section, we show experiments with the proposed vision-based end-to-end autonomous driving framework in both sim-to-real environments and a full-scale autonomous vehicle. We start by introducing the hardware platform and data collection, followed by implementation details of the proposed model. We then demonstrate extensive analysis in the sim-to-real environment VISTA \cite{amini2021vista}. Finally, we showcase results with real-car deployment.

\subsection{Hardware Setup and Real-world Data Collection}
We deploy our models onboard a full-scale autonomous vehicle (2019 Lexus RX 450H) equipped with a NVIDIA 2080Ti GPU and an AMD Ryzen 7 3800X 8-Core Processor. We use a RGB camera BFS-PGE-23S3C-CS as the primary perception sensor, which runs in 30Hz, with a resolution of 960$\times$600, and has 130$^\circ$ horizontal field-of-view. Other onboard sensors include inertial measurement sensor (IMUs) and wheel encoders to measure steering feedback and odometry. Also, we use a differential global positioning system (dGPS) for evaluation purpose. To run the data-driven simulation VISTA \cite{amini2021vista}, we collect real-world data from a wide-range of environments, including different time of day, weather conditions, and seasons of a year. The entire dataset consists of roughly 2 hour of driving data, which is further augmented with our training dataset generation pipeline using VISTA.

\subsection{Synthetic Training Dataset Generation}
\label{ssec:dset_generation}
We train our model with guided policy learning, which has been shown to improve effectiveness for direct model transfer to real-car deployment. The data generation process follows (a) in VISTA, randomly initializing both ego- and ado-car with different configurations like relative poses, geographical locations associated with the real dataset, appearance of the vehicle, etc (b) running an optimal controller with access to privileged information to steer the ego-vehicle and collect ground-truth control outputs with corresponding states, and (c) collecting RGB images at viewpoints along the trajectories. We choose nonlinear Model Predictive Control (NMPC) as the privileged (nominal) controller. While NMPC is usually computationally expensive and hard to solve, it is tractable offline and, with jerk $u_{jerk}$ and steering acceleration $u_{steer}$ as controls, provides smooth acceleration $a$ and steering rate $w$, which is used as learning targets in BarrierNet. Vehicle dynamics of NMPC and BarrierNet \eqref{eqn:affine} are defined with respect to a reference trajectory \cite{Rucco2015}. It measures the along-trajectory distance $s\in\mathbb{R}$ and the lateral distance $d\in\mathbb{R}$ of the vehicle Center of Gravity (CoG) with respect to the closest point on the reference trajectory, 

{\small\begin{equation} \label{eqn:vehicle}
   \underbrace{\left[
\begin{array}[c]{c}
    \dot s\\
    \dot d\\
    \dot \mu\\
    \dot v\\
    \dot a\\
    \dot \delta\\
    \dot \omega
\end{array}
\right]}_{\dot {\bm x}}
=
\underbrace{\left[
\begin{array}
[c]{c}%
    \frac{v\cos(\mu + \beta)}{1 - d\kappa}\\
    v\sin(\mu + \beta)\\
    \frac{v}{l_r}\sin\beta - \kappa\frac{v\cos(\mu + \beta)}{1 - d\kappa}\\
    a\\
    0\\
    \omega\\
    0
\end{array}
\right]}_{f(\bm x)}
+
\underbrace{\left[
\begin{array}[c]{cc}%
    0 & 0\\
    0 & 0\\
    0 & 0\\
    0 & 0\\
    1 & 0\\
    0 & 0\\
    0 & 1
\end{array}
\right]}_{g(\bm x)}
\underbrace{\left[
\begin{array}[c]{c}%
    u_{jerk}\\
    u_{steer}
\end{array}
\right]}_{\bm u},
\vspace{-2pt}
\end{equation}
}

\begin{figure}[t]
	\centering
	\vspace{-4mm}
	\includegraphics[scale=0.28]{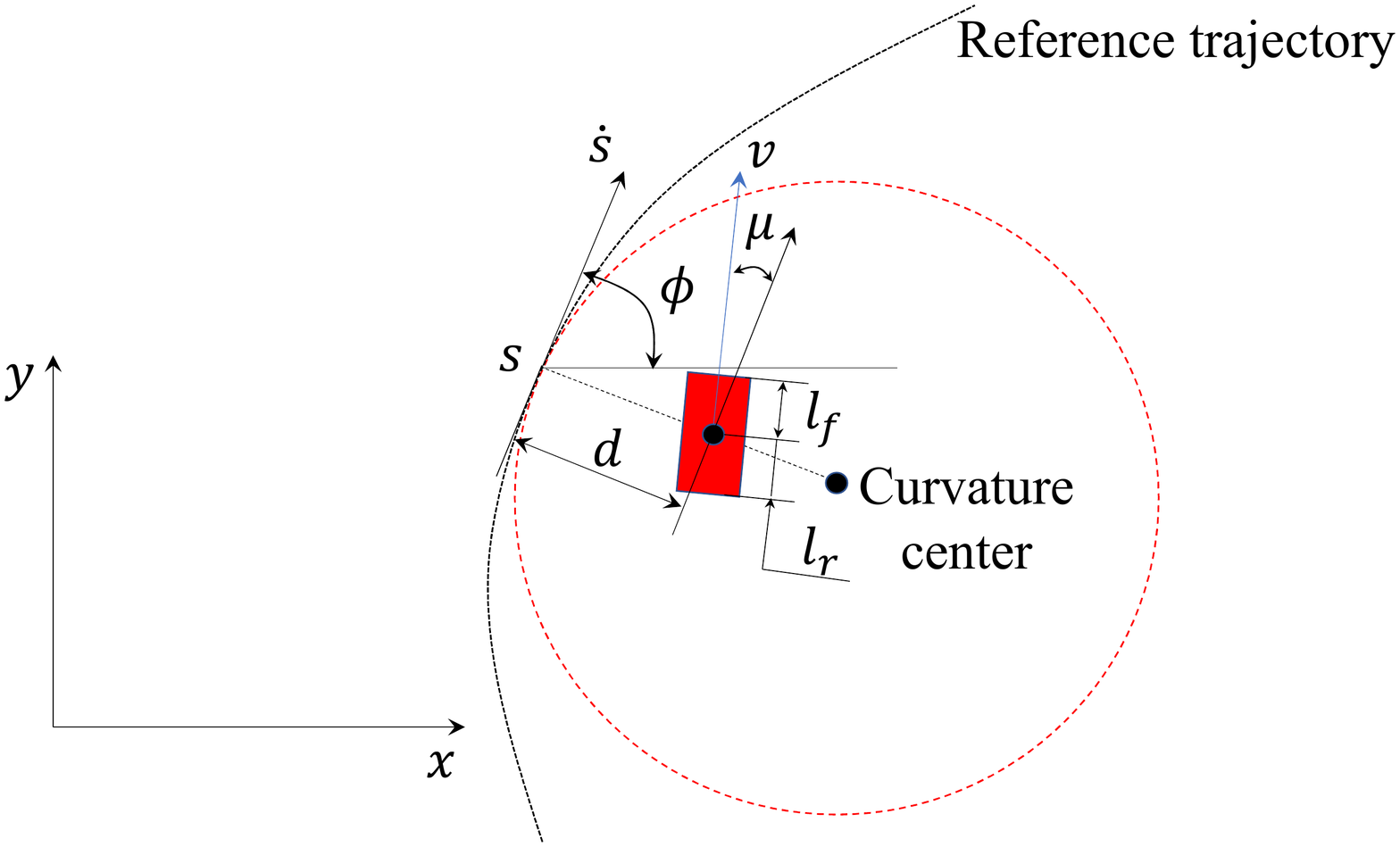}
	\vspace{-9mm}
	\caption{Coordinates of ego w.r.t a reference trajectory. }	
	\vspace{-4mm}
	\label{fig:frame}
\end{figure}

\noindent where $\mu$ is the vehicle local heading error determined by the difference of the global vehicle heading $\theta\in\mathbb{R}$ and the tangent angle $\phi\in\mathbb{R}$ of the closest point on the reference trajectory (i.e., $\theta = \phi + \mu$) as shown in Fig. \ref{fig:frame}; $v$, $a$ denote the vehicle linear speed and acceleration; $\delta$, $\omega$ denote the steering angle and steering rate, respectively; $\kappa$ is the curvature of the reference trajectory at the closest point; $l_r$ is the length of the vehicle from the tail to the CoG; and $u_{jerk}$, $u_{steer}$ denote the two control inputs for jerk and steering acceleration (in the nominal controller). $\beta = \arctan\left(\frac{l_r}{l_r + l_f}\tan\delta\right)$, where $l_f$ is the length of the vehicle from the head to the CoG.
We set the receding horizon of the NMPC to 20 time steps during data sampling, and it is implemented in a virtual simulation environment in MATLAB. We augment the real-world dataset using VISTA and NMPC with synthetic obstacle avoidance and lane following data. In total, the training dataset has around 400k images.


\noindent \textbf{Implementation Details.} Based on the aforementioned general framework in Sec. \ref{ssec:dcbfs}, we specifically define dCBFs for lane following as $b_{lf}^{left}=d_{lf}-d$ and $b_{lf}^{right}=d_{lf}+d$ (besides the disk covering lane following approach shown in the left case of Fig. \ref{fig:cover} as we only study single obstacle avoidance), and for obstacle avoidance as $b_{obs}=\Delta s^2 + (d-d_{obs})^2 - r_D^2$, where $d_{lf}$ is a preset bound, $\Delta s$ is relative progress between ego-car and obstacle, $r_D$ is disk size, and $d_{obs}$ is the lateral displacement from lane center of the obstacle. We compute Lie derivative to construct dCBF constraints in QP with vehicle dynamics mentioned above. Overall, the model takes in a front-view image, infers speed $v$ and steering angle $\delta$ to compute reference control with derivative and state $(d, d_{obs}, \mu, \Delta s)$, predicts dCBFs parameters, and obtains final control $(a, \omega)$ by solving QP with dCBFs. The learning supervision includes Mean Squared Error loss on $v, \delta, d, d_{obs}, \mu, \Delta s, a, \omega$. We bound the derivative of $v,\delta$ (reference control) to stabilize learning. We cap loss on $\Delta s, d_{obs}$ when the obstacle is absent or too far away, to ensure states can be reasonably predicted.

\subsection{Evaluation In Sim-to-Real Environments}
Open-loop control error (i.e., difference between predicted and ground-truth control) has been shown to be a poor indicator to evaluate the performance of a driving policy since it only measures error around ground-truth trajectories and ignores accumulated errors that gradually drift the vehicle to out-of-distribution regions. Hereby, we presents closed-loop testing results in the sim-to-real environment VISTA \cite{amini2021vista}.

\begin{figure}[t]
	\centering
	\includegraphics[scale=0.29]{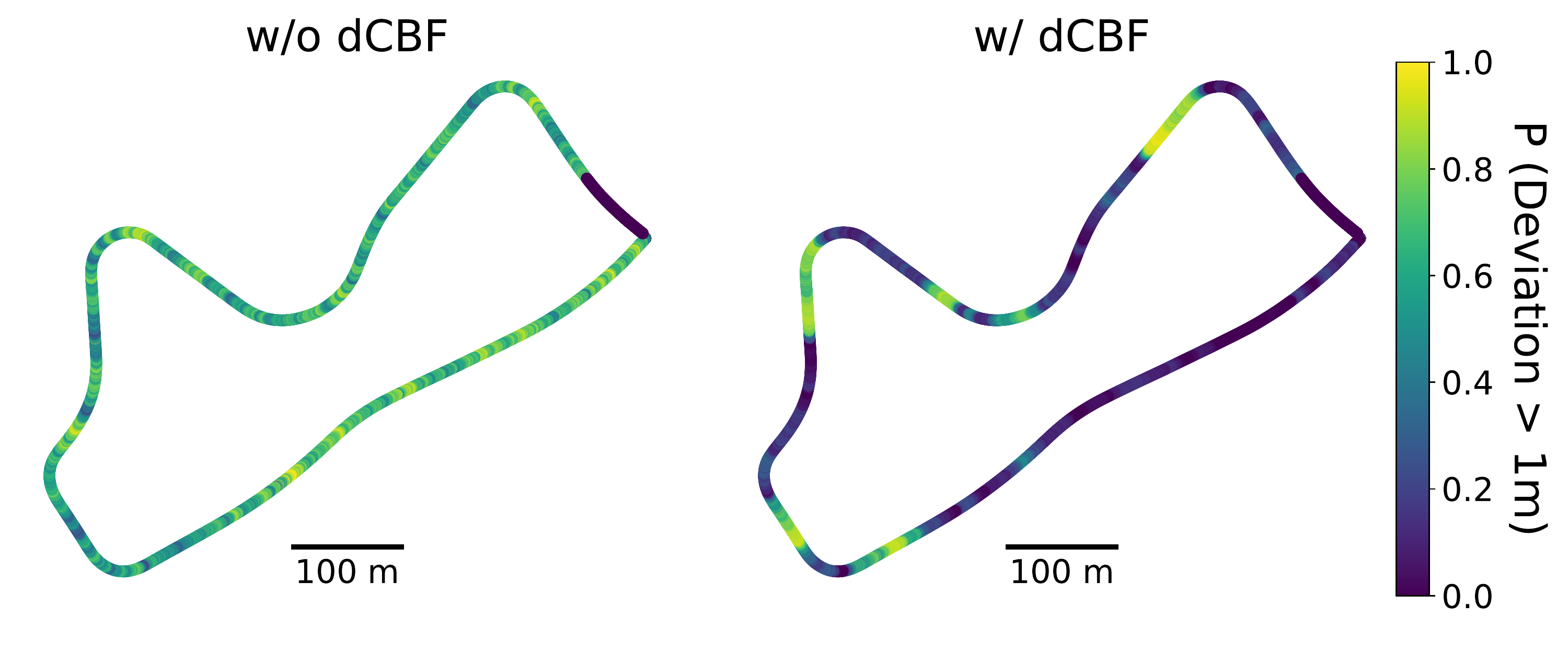}
	\caption{Lane following probabilistic comparisons of deviation from the lane center in a BarrierNet with/without lane keeping CBFs.}	
	\label{fig:loop}
\end{figure}

\noindent \textbf{Lane Keeping As Safety Constraints.} 
In Fig. \ref{fig:loop}, we show the probability of ego-vehicle deviating away from the lane center for larger than 1m. We run 1000 episodes with maximal 200 steps if not crashed (off-lane more than 2m) prematurely. In each episode, the vehicle is randomly initialized at a point in the trace and we compute average deviation at every point to ensure sufficiently large sample size for the statistics. The model with lane keeping CBFs achieves significantly better performance since they can encourage the autonomous vehicle to stay close to the lane center by decreasing the boundary values due to the Lyapunov property of CBFs.

\begin{table}
	\caption{Crash rate and clearance with/without BarrierNet, using or not using ground truth obstacle information.}
	\label{tab:comp}
	\centering
	\small
	\begin{tabular}
	{ c|c|c } 
	    \toprule
		Method    & Crash Rate $\downarrow$ & Min. Clearance (m) $\uparrow$ \\\midrule
		w/o dCBF & 0.53 & 0.43 \\
		w/ dCBF & 0.28 & 0.55 \\
		w/ dCBF (with gt) & 0.03 & 0.61 \\
		\bottomrule
	\end{tabular}
\end{table}

\begin{figure}[t]
	\centering
	\includegraphics[scale=0.35]{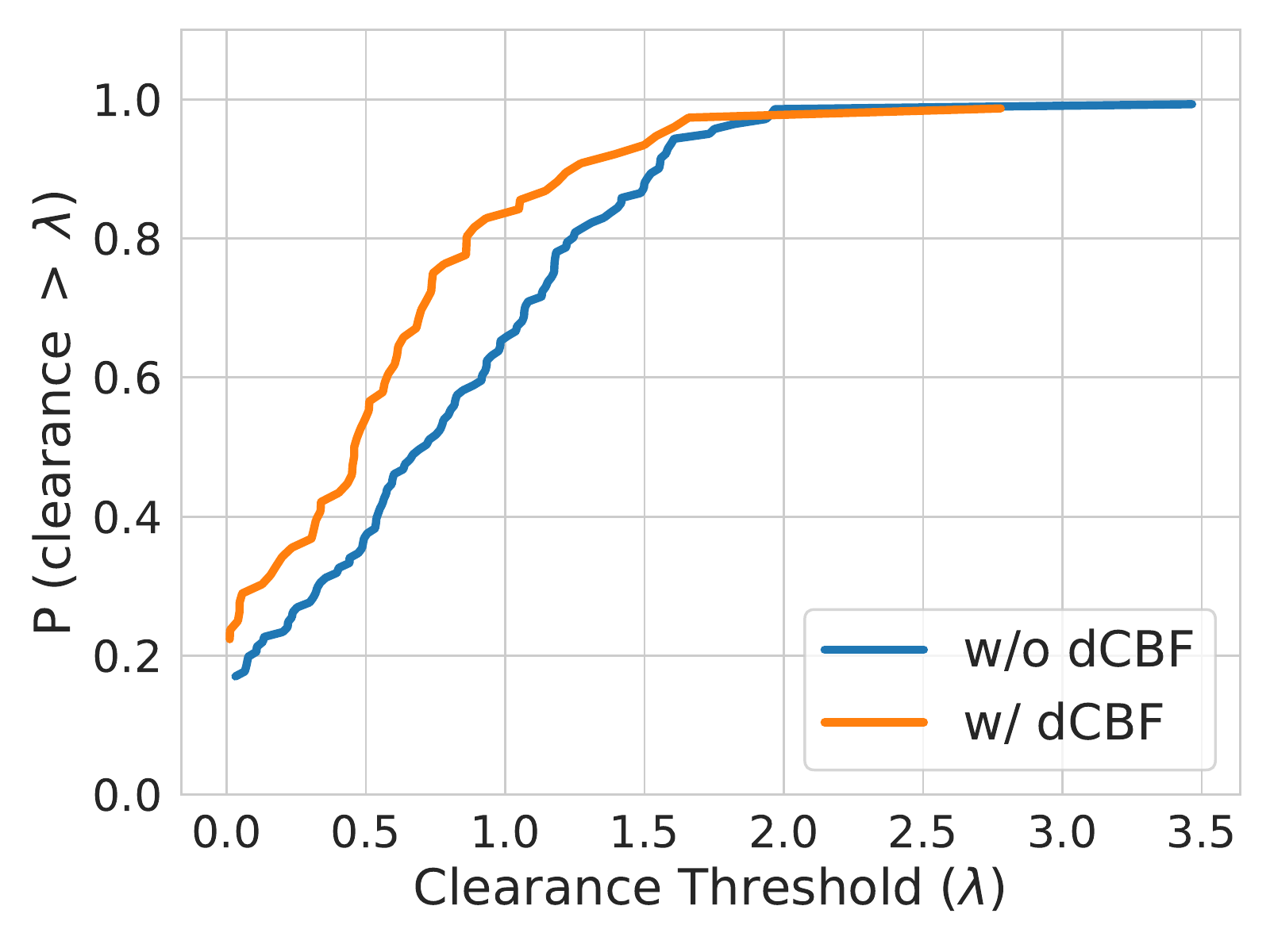}
	\caption{Line plot for clearance in obstacle avoidance with/without BarrierNet.}	
	\label{fig:obs_clear}
\end{figure}

\noindent \textbf{Obstacle Avoidance.} 
In Table \ref{tab:comp}, we show crash rate and minimal clearance of models with or without BarrierNet and with or without access to ground-truth states. Minimal clearance is computed as the closest distance between polygons of ego- and ado-car within an episode. The introduction of obstacle avoidance dCBFs significantly reduces crash rate and increases clearance. The remaining failures mainly come from the imprecise or even erroneous state and obstacle information inferred from the front-view camera only. With access to ground-truth information (an ideal state estimator), the crash rate is close to yet not zero. This might be due to misaligned dynamics and inter-sampling effects of CBFs which have been extensively studied in CBFs \cite{taylor2020learning, Xiao2021TAC1}. To look deeper into how BarrierNet improves safety distance, we plot the distribution of clearance larger than a varying threshold among all time steps in Fig. \ref{fig:obs_clear}, where larger area under the curve indicates better safety.



\begin{figure}[t]
	\centering
	\includegraphics[scale=0.4]{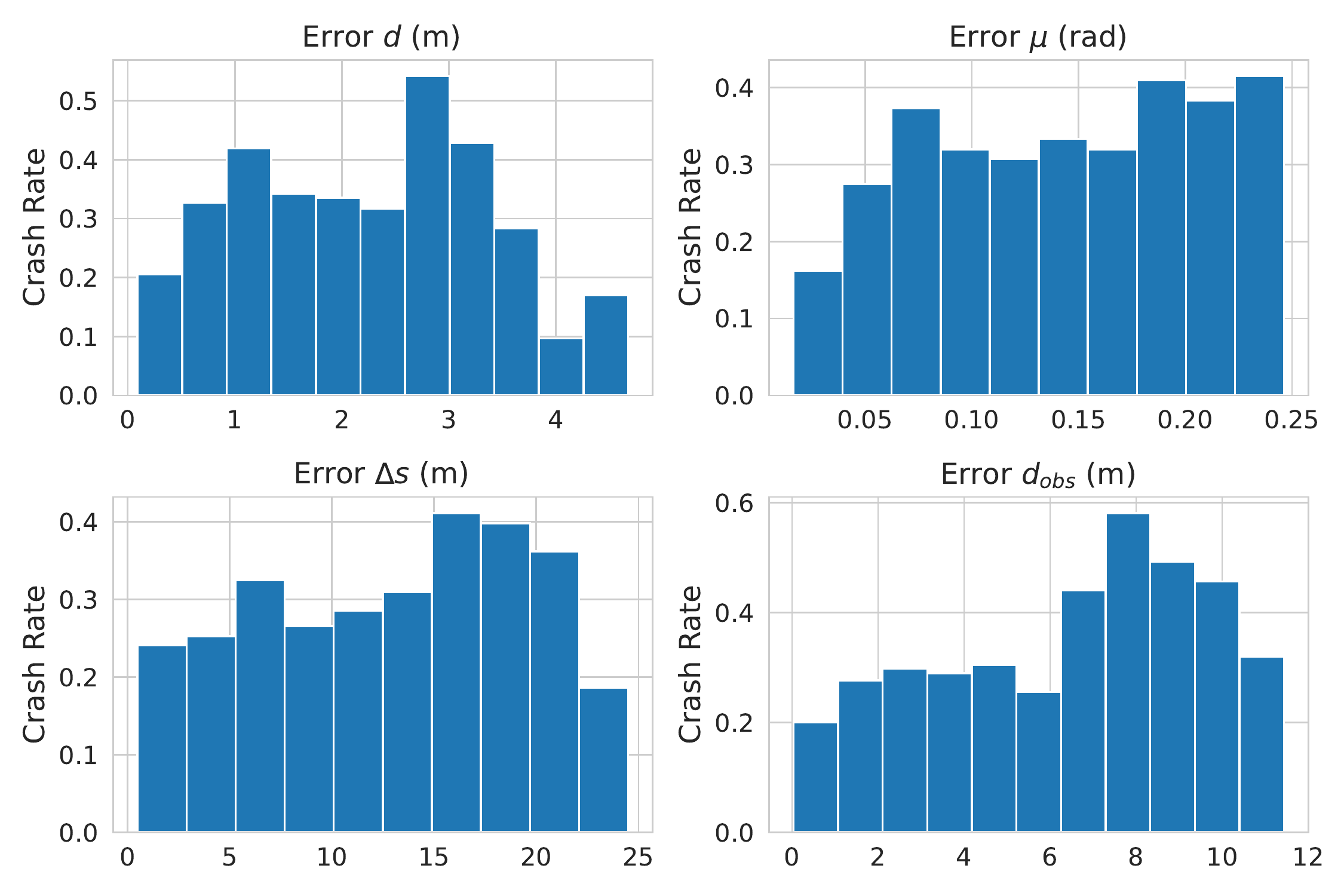}
	\caption{Statistics for crash rate in obstacle avoidance under different levels of prediction error from front vision view.}	
	\label{fig:obs_clear_pred}
\end{figure}

Furthermore, in Fig. \ref{fig:obs_clear_pred}, we investigate how imperfect state estimation introduces error in dCBFs and affects crash rate. We show the four output prediction from the state estimation module, including deviation from lane center of ego-car $d$, local heading error $\mu$ with respect to road curvature, relative progress along the road between ego-car and the obstacle $\Delta s$, and lateral displacement of obstacle from lane center $d_{obs}$. As expected, the overall trend shows increasing failure with larger state estimation error. The performance drops drastically with large $d_{obs}$ since roughly it indicates whether the obstacle is at the left or right with respect to the ego-car and thus has great influence within obstacle avoidance dCBFs. Note that bins at large error can have fewer samples (e.g., the tail in histogram of $d$) and may lead to high variance in the estimator. We still keep the results for completeness. The results highlight the importance of handling state estimation error and suggest future research on uncertainty calibration, which will be the focus of our continued effort. 




\begin{figure}[t]
	\centering
	\includegraphics[scale=0.5]{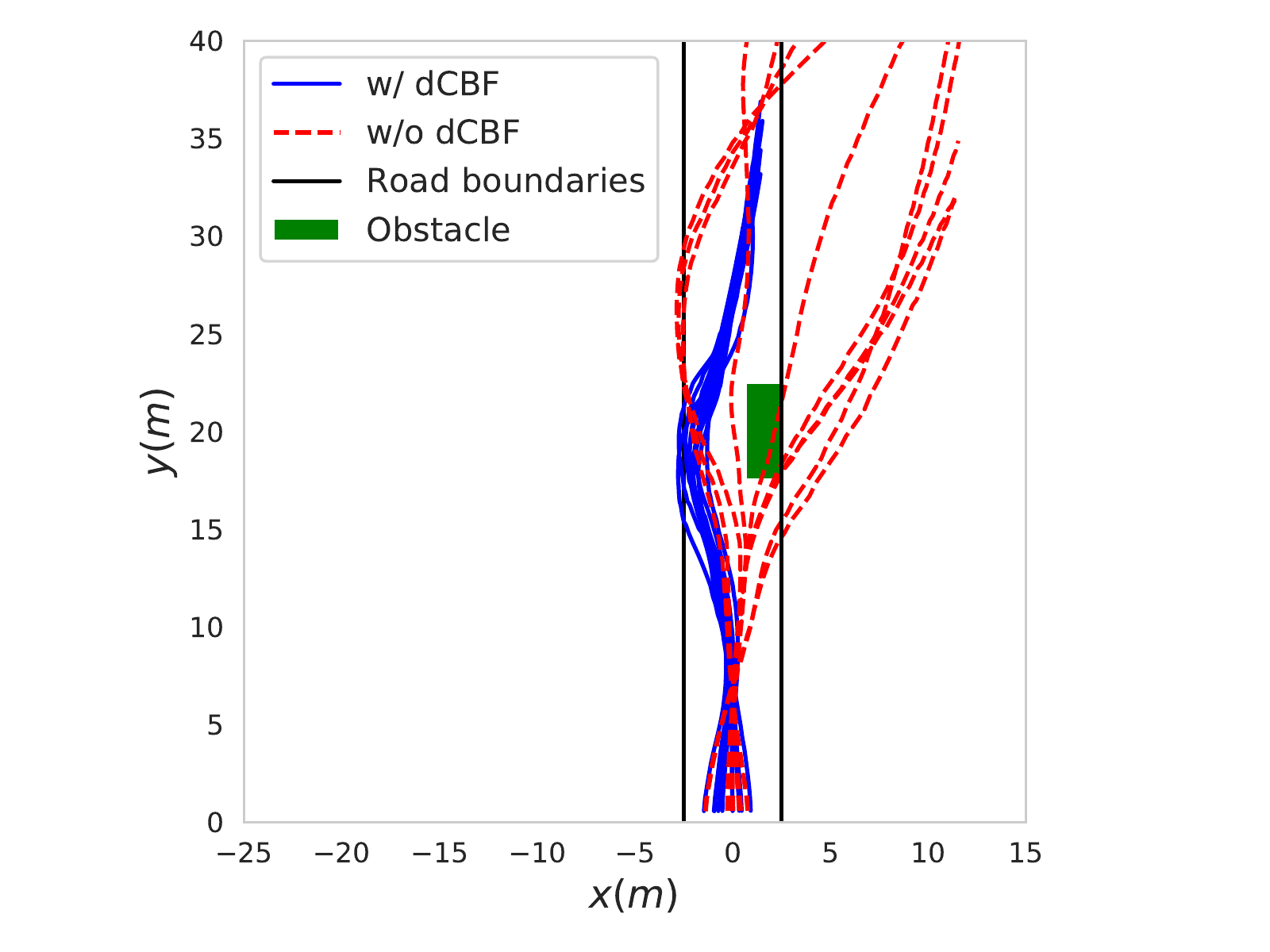}
	\vspace{-4mm}
	\caption{Vehicle trajectories in obstacle avoidance with/without BarrierNet in VISTA.}	
	\label{fig:obs_traj}
\end{figure}

\begin{figure}[t]
	\centering
    \begin{subfigure}{0.24\textwidth}
        \centering
        \includegraphics[width=\textwidth]{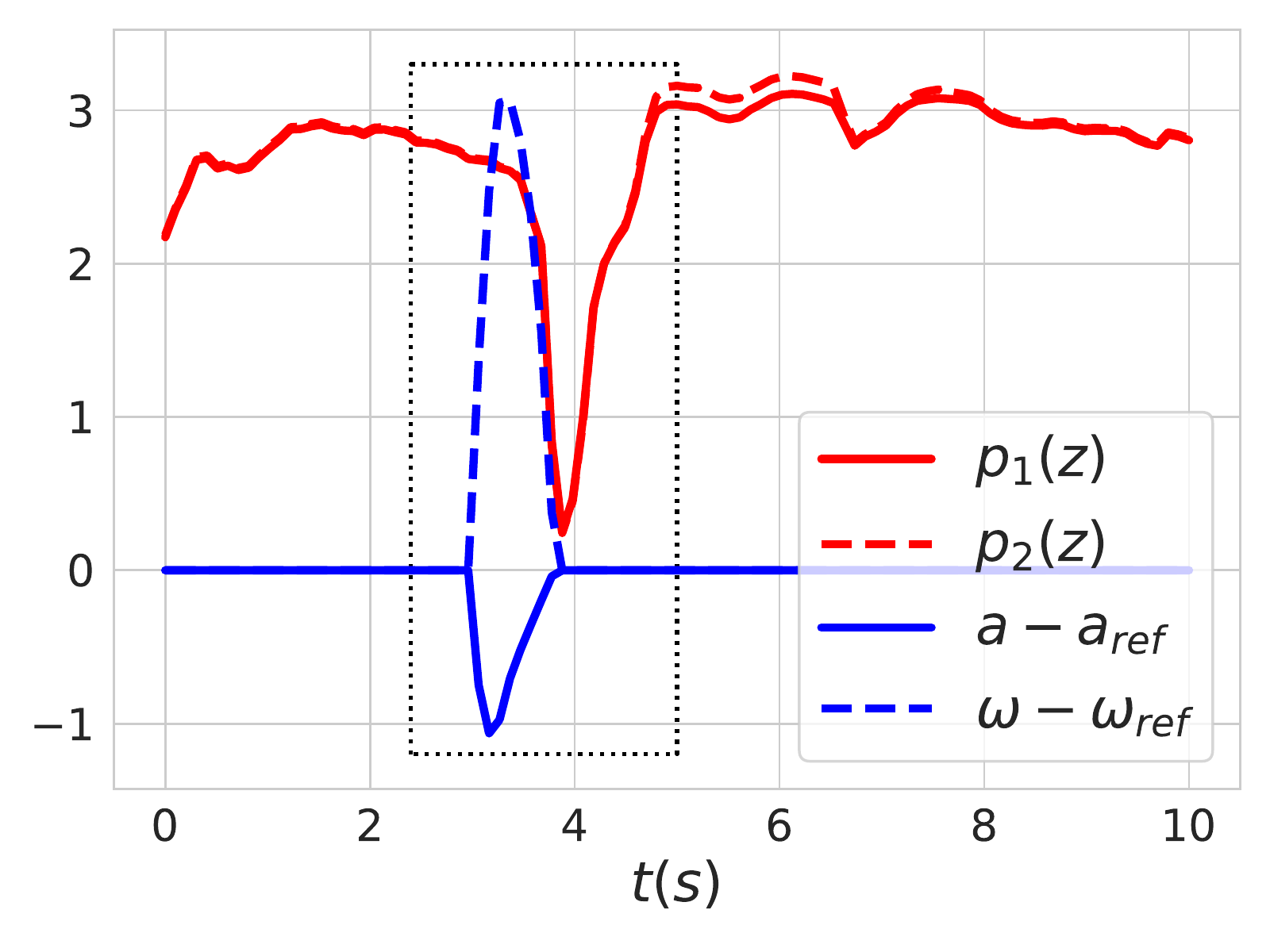}
    \end{subfigure}
    \begin{subfigure}{0.24\textwidth}
        \centering
        \includegraphics[width=\textwidth]{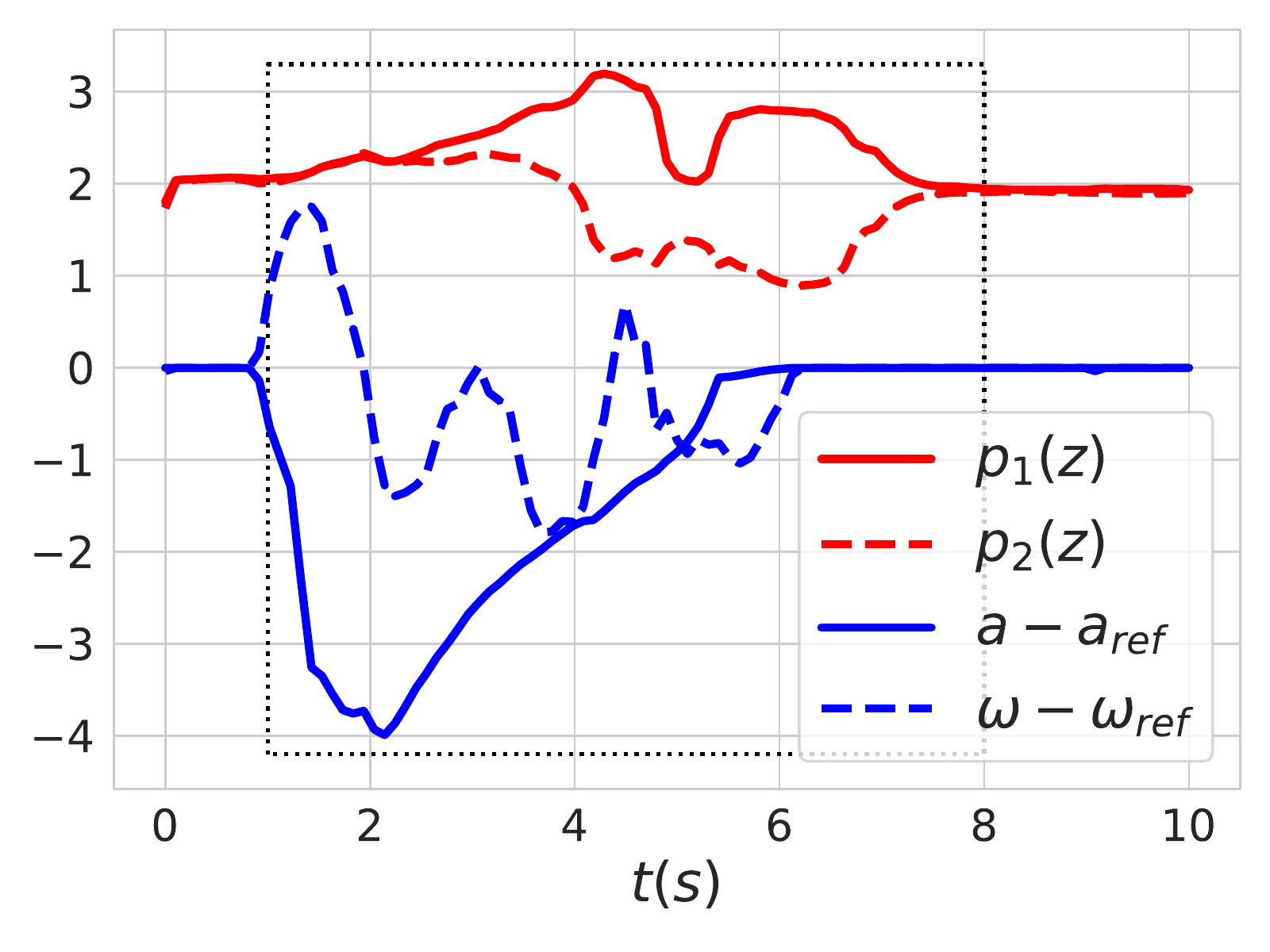}
    \end{subfigure}
	\caption{Penalty $p_1(\bm z), p_2(\bm z)$ variation in a dCBF (BarrierNet) when approaching an obstacle under two (different) trained BarrierNets. The relative degree of the safety constraint is two, and thus we have two CBF parameters in one CBF. The segments inside the dotted boxes denote intervals when the ego vehicle is near the obstacle. The box sizes are different as the ego has different speeds when passing the obstacle.}	
	\label{fig:obs_penalties}
\end{figure}

\begin{figure*}[t]
	\centering
	\includegraphics[scale=0.4]{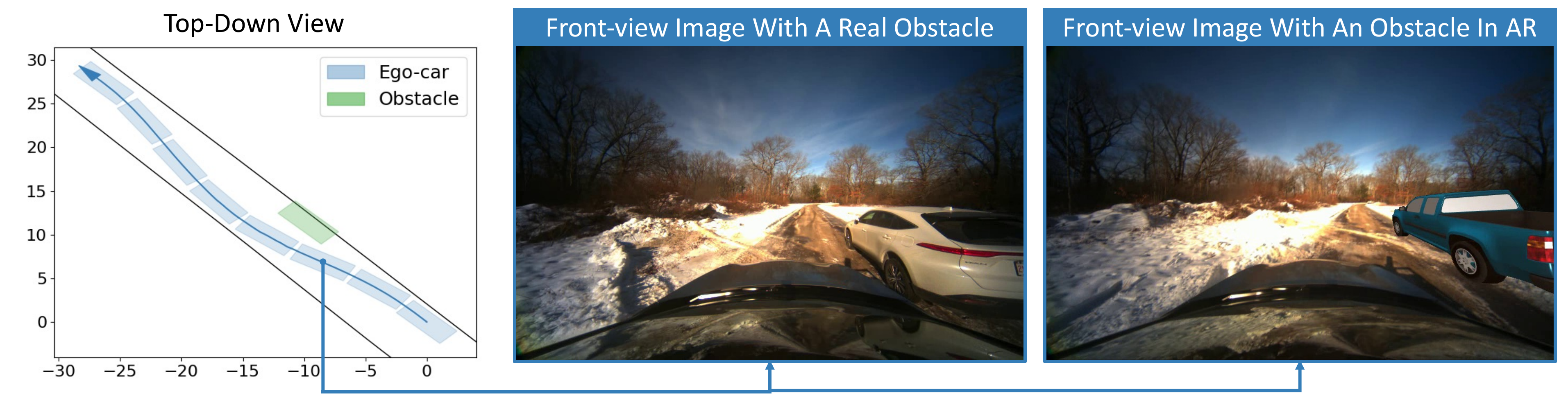}
	\caption{An illustration of real-car experiments.}	
	\label{fig:exp_illustration}
\end{figure*}

\noindent \textbf{BarrierNet Provides Safe Maneuvers.} 
In Fig. \ref{fig:obs_traj}, we benchmarked for different learning systems, with and without BarrierNet, to provide driving trajectories under the same configuration except for arbitrary initial pose on a road. We show 10 variants of initial states for each model. It can be observed that trajectories from the two models mostly align with each other in the beginning as the ego-vehicle starts with different lateral displacement from the lane center and tries to recover. Then, the two set of trajectories diverges while approaching the obstacle. This is the consequence of correction from the activated dCBFs over the reference unsafe control. With BarrierNet, the safety is guaranteed.

\noindent \textbf{BarrierNet With Different Profiles.} We also notice that the BarrierNet may learn different CBF parameters when the ego vehicle approaches an obstacle. In Fig. \ref{fig:obs_penalties}, we present two possible variations of penalty functions $p_1(\bm z), p_2(\bm z)$ when the ego vehicle is around an obstacle. The penalty functions $p_1(\bm z), p_2(\bm z)$ adapt to the obstacle when the ego vehicle is close to an obstacle, and they recover to some values when the ego leaves the obstacle. This shows the flexibility of the BarrierNet. Another observation is that the outputs of the BarrierNet tend to deviate from the reference controls (from the previous LSTM layer) when the ego vehicle is close to the obstacle. This shows the safety guarantee property of CBFs. In order to avoid this deviation, we need to improve the learned model with better reference controls and CBF parameters.


\subsection{Physical Autonomous Car Experiments}
To verify the effectiveness of the proposed vision-based end-to-end framework with dCBFs, we deploy the trained models on a full-scale autonomous driving car. The experiments are conducted in a test site with rural road type. We majorly test the algorithm with augmented reality (AR) and only perform minimal experiment with real-car obstacle for safety reasons. We use a pre-collected map of the test site and vehicle pose from the differential GPS (dGPS) to place virtual obstacles in the front of the ego-vehicle on road with AR. Note that the tested models are still using vision inputs only to steer the autonomous vehicle without any access to ground-truth state. Fig. \ref{fig:exp_illustration} is an illustration of the real-car experimental setup. 
Another thing worth mentioning is that the scene is covered with snow at the time we conducted real-car experiment. The icy road surface at the track and heavy snow at the side of the road introduce tire slippage and pose additional challenges to our self-driving system. Also, the reflection of sunlight on the ice makes it hard to recognize road boundaries even from human judgement.
With high-precision dGPS in the site (covariance $<$ 1cm), we provides qualitative analysis with side-by-side comparison between models with and without BarrierNet.

\begin{figure}[thpb]
	\centering
    \begin{subfigure}{0.24\textwidth}
        \centering
        \includegraphics[width=\textwidth]{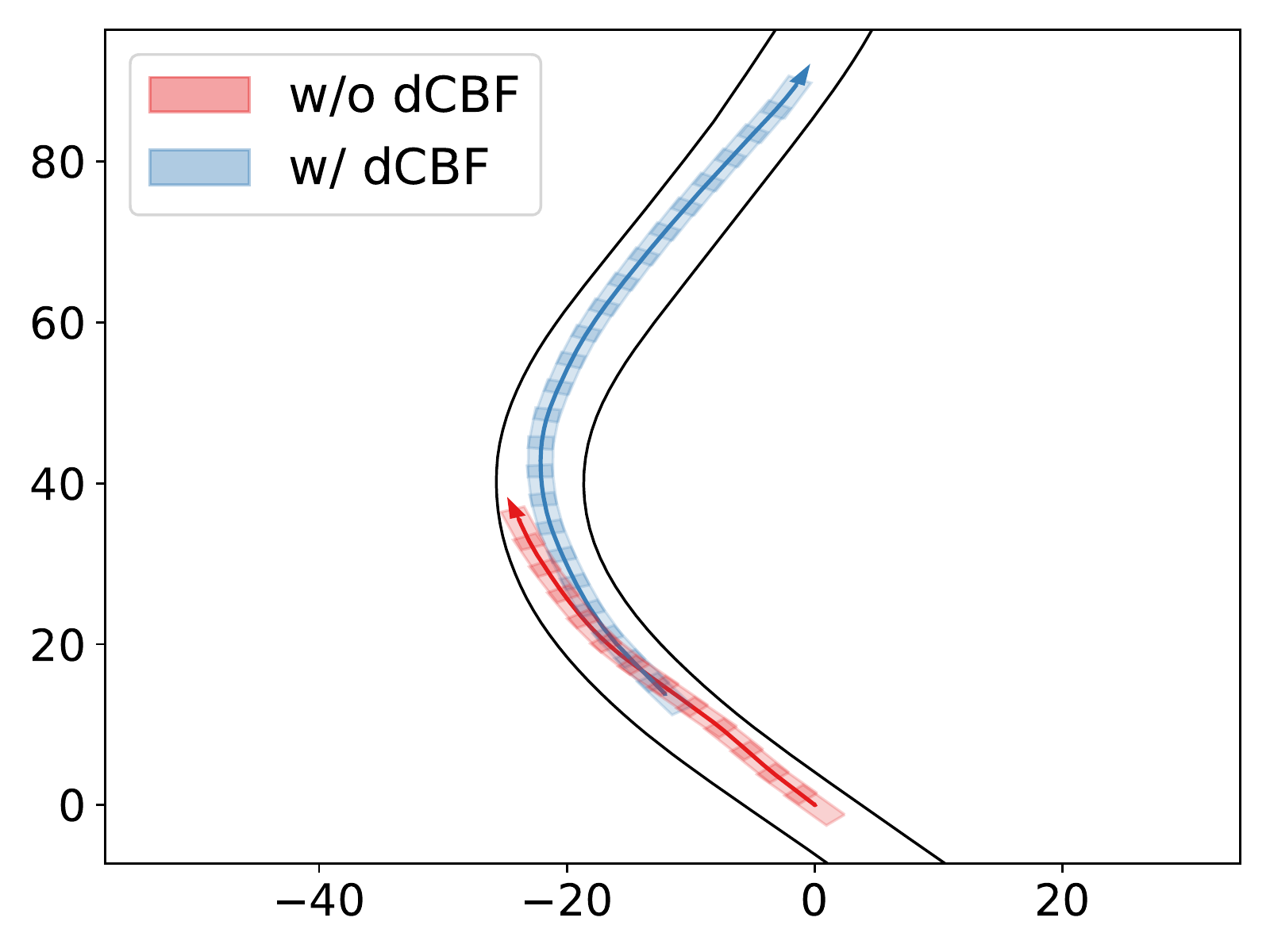}
    \end{subfigure}
    \begin{subfigure}{0.24\textwidth}
        \centering
        \includegraphics[width=\textwidth]{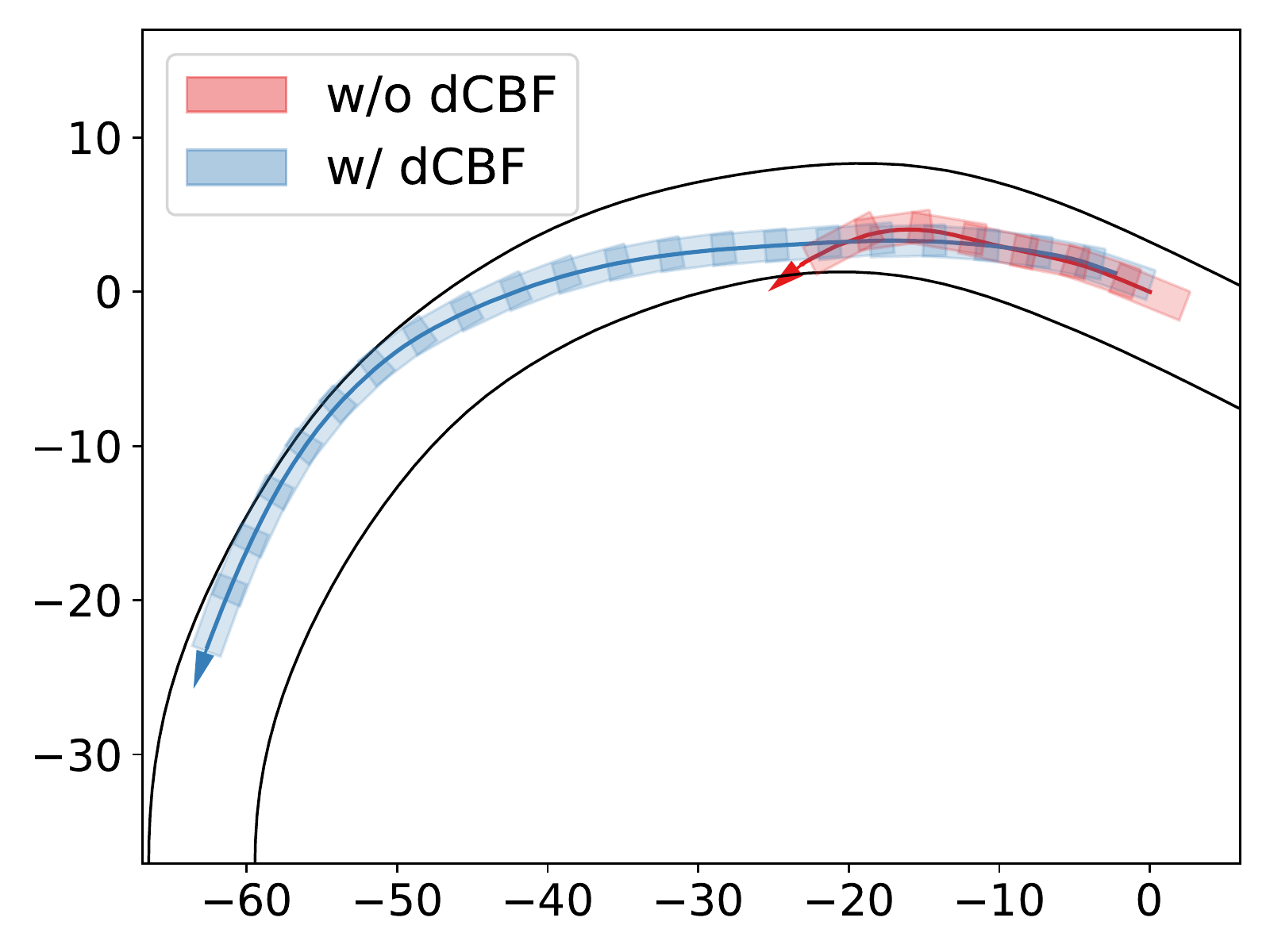}
    \end{subfigure}
	\caption{Two cases of experimental vehicle trajectories in obstacle avoidance with/without lane keeping CBFs in the BarrierNet. Tire slipping happens on the icy road.}	
	\label{fig:real_car_lane_following}
\end{figure}

\noindent \textbf{BarrierNet In Challenging Sharp Turns.} 
In Fig. \ref{fig:real_car_lane_following}, we demonstrate driving trajectories of BarrierNet with and without lane keeping CBFs in sharp left and right turns. We show the footprint of vehicle through time and indicate forward direction with arrows. Without lane keeping CBFs (red), the car is more prone to get off-road, while roughly correct estimates of deviation from lane center ($d$) imposes an additional layer of safety with lane keeping CBFs (blue).

\begin{figure}[thpb]
	\centering
    \begin{subfigure}{0.24\textwidth}
        \centering
        \includegraphics[width=\textwidth]{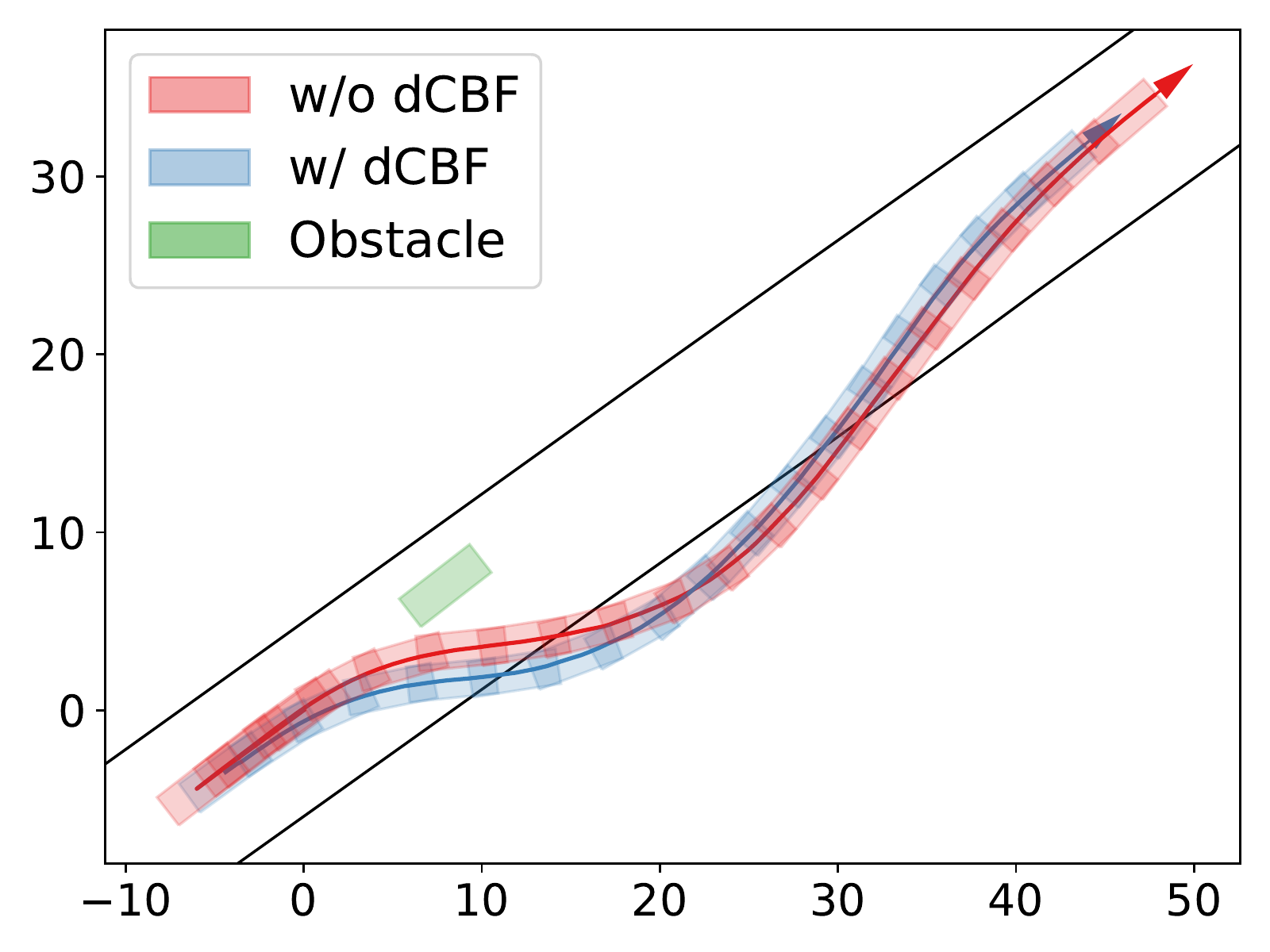}
    \end{subfigure}
    \begin{subfigure}{0.24\textwidth}
        \centering
        \includegraphics[width=\textwidth]{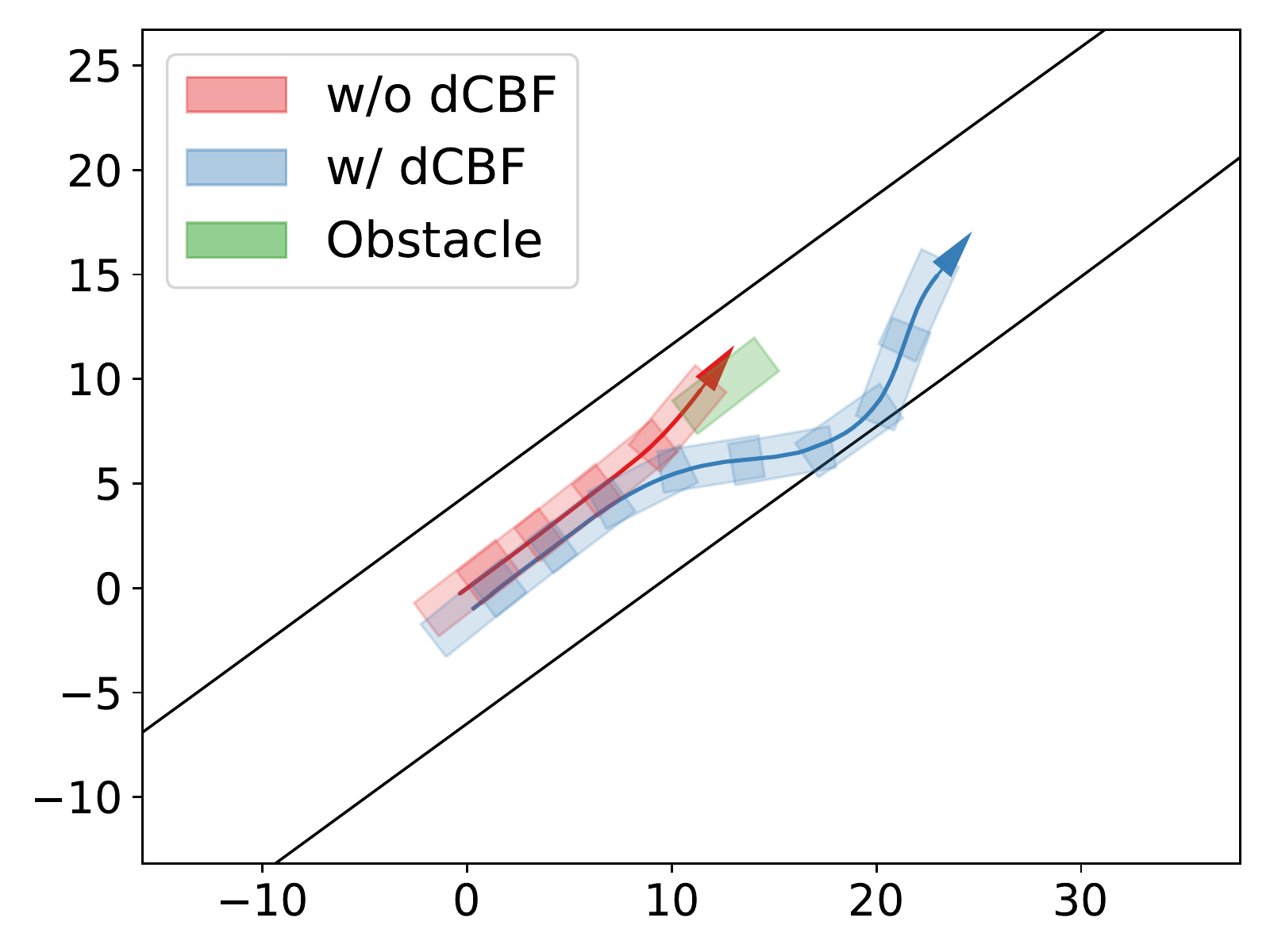}
    \end{subfigure}
	\caption{Two cases of experimental vehicle trajectories in obstacle avoidance with/without BarrierNet. In the left case, the heavy snow by the road is preventing the vehicle from getting back to the road due to tire slipping, and thus the vehicle recovers slowly even when the steering wheel is at its left limit.}	
	\label{fig:real_car_obs_avoidance}
\end{figure}

\noindent \textbf{Obstacle Avoidance In Real World.} 
We also did experiments on the autonomous car in obstacle avoidance, as shown in Fig. \ref{fig:real_car_obs_avoidance}. The first example (left) demonstrates that with reasonable reference control (both models successfully avoid the obstacle), the model with obstacle avoidance dCBFs (blue) creates more clearance to achieve better safety. The second example (right) highlights the effectiveness of BarrierNet (blue) when the reference control (red) fails to avoid the front car and requires correction from activated dCBF constraints. 


\section{Conclusion} 
\label{sec:conclusion}
We proposed an end-to-end learning framework for obtaining safety-guaranteed perception-based autonomous driving agents. Our method is constructed by combining a deep neural model with a differentiable higher-order control barrier function to ensure safe lane-keeping and obstacle avoidance. We showed how various modules of our pipeline can contribute to the understanding of the learning system's behavior while driving. Our sim-to-real and real-world experiments demonstrated the effectiveness of our approach in many driving scenarios where we could strictly reduce the probability of crash and interventions, when our pipeline is activated.

We hope that our method can inspire future research on endowing real-world robot learning schemes with fundamental control theory modules to enhance interpretability, robustness and safety.

\section*{Acknowledgments} 
This work was partially supported by Capgemini Engineering. This research was also sponsored by the United States Air Force Research Laboratory and the United States Air Force Artificial Intelligence Accelerator and was accomplished under Cooperative Agreement Number FA8750-19-2-1000. The views and conclusions contained in this document are those of the authors and should not be interpreted as representing the official policies, either expressed or implied, of the United States Air Force or the U.S. Government. The U.S. Government is authorized to reproduce and distribute reprints for Government purposes notwithstanding any copyright notation herein. 

\bibliographystyle{plainnat}
\bibliography{MCBF}

\end{document}